\documentclass{article}

\PassOptionsToPackage{numbers, compress}{natbib}

\usepackage[main, final]{neurips_2026}

\usepackage[utf8]{inputenc} 
\usepackage[T1]{fontenc}    
\usepackage{url}            
\usepackage{booktabs}       
\usepackage{amsfonts}       
\usepackage{amsmath}
\usepackage{nicefrac}       
\usepackage{microtype}      
\usepackage{xcolor}         
\definecolor{citeblue}{RGB}{48,111,186}
\usepackage[pagebackref=false,breaklinks=true,colorlinks=true,citecolor=citeblue,bookmarks=false]{hyperref}
\usepackage{cleveref}  
\usepackage{multirow}
\usepackage{multicol}
\usepackage{xspace}
\usepackage{graphicx}
\usepackage{pifont}
\usepackage{wrapfig}

\crefname{section}{Sec.}{Secs.}
\Crefname{section}{Section}{Sections}
\crefname{table}{Tab.}{Tabs.}
\Crefname{table}{Table}{Tables}
\crefname{figure}{Fig.}{Figs.}
\Crefname{figure}{Figure}{Figures}
\crefname{equation}{Eq.}{Eqs.}
\Crefname{equation}{Equation}{Equations}
\crefname{appendix}{Appendix}{Appendix}
\newcommand{\eg}{\textit{e.g.}\xspace}
\newcommand{\ie}{\textit{i.e.}\xspace}

\newcommand{\vs}{\textit{v.s.}\xspace}

\newcommand{\method}{LongSCOP\xspace}
\newcommand{\bench}{LVO-Bench\xspace}

\title{LongSCOP: Semantically Consistent\\ Long Video Outpainting}

\author{
Yu Tang$^{13*}$, Zhiyuan You$^{23*}$, Zhixiao Wang$^1$, Fanghua Yu$^4$, Qingyu Zhang$^1$, \\\textbf{ Xiang Yin$^{3}$, Chao Dong$^{35\dag}$, Jinjin Gu$^{6\dag}$}\\
$^1$Nanjing University\quad $^2$The Chinese University of Hong Kong\\
$^3$Shenzhen Institutes of Advanced Technology, Chinese Academy of Sciences\\
$^4$The University of Hong Kong\quad
$^5$Shenzhen University of Advanced Technology\\
$^6$INSAIT, Sofia University ``St. Kliment Ohridski''\\
{* Equal Contribution  \quad $^\dag$ Corresponding Author}
}

\begin{document}

\maketitle
\vspace{-20pt}

\begin{abstract}

Video out-painting is an important problem, yet existing research focuses primarily on short clips, leaving long video out-painting largely underexplored.
Extending video out-painting to long videos poses substantial challenges.
A natural solution is iterative out-painting, \ie, generating videos chunk by chunk, but this introduces two major issues: generation instability or even failure caused by inter-chunk inconsistencies and error accumulation, and semantic drift over long temporal horizons.
In this paper, we propose \textbf{LongSCOP}, short for \textbf{S}emantically \textbf{C}onsistent long Video \textbf{O}ut-\textbf{P}ainting.
First, we introduce a training recipe with three components for smooth and stable long-horizon generation: self-forcing training, latent calibration, and future references.
Beyond generation stability, we recognize that the rich temporal context of long videos provides valuable references for maintaining semantic consistency. We therefore develop a VLM-based agent that automatically selects the most informative reference frames from the entire video to maintain consistent subjects, scenes, and backgrounds throughout long-horizon generation.
We also construct a high-quality training dataset of 62K video clips.
To facilitate research on this underexplored problem, we introduce \textbf{LVO-Bench}, a benchmark curated by human experts to assess long-horizon stability and semantic consistency.
Extensive experiments demonstrate that our method outperforms existing approaches in both generation stability and semantic coherence.

\end{abstract}

\vspace{-12pt}
\section{Introduction}\label{sec:intro}
\vspace{-5pt}

Long video out-painting is an important yet underexplored problem. 
Video out-painting aims to expand the spatial field of view of a video beyond its originally observed boundaries~\cite{wang2024beyouroutpainter}, and has many applications~\cite{yu2025unboxed, dehan2022videooutpainting}. 
Benefiting from the rapid progress of video generation models~\cite{lin2024open,wan2025wan}, recent video out-painting methods have achieved encouraging results in synthesizing visually realistic and temporally smooth content~\cite{zhong2025outdreamer,chen2025infinitecanvas}. 
Meanwhile, extending out-painting from short clips (\eg, a few seconds) to substantially longer videos (\eg, 20 seconds or more) is increasingly important in real-world usage. 
Such long-form content has become a dominant format on modern video platforms~\cite{elmoghany2025survey}, and it usually contains richer narratives and more dynamic scene transitions compared to short clips; thus, it is more valuable for practical applications~\cite{yu2025unboxed}.

Despite its practical importance, long video out-painting remains highly challenging. 
A direct solution would be to process an entire long video jointly. 
However, due to the high computational and memory cost of video generation models, handling long sequences in a single pass is often impractical. 
A more natural and scalable alternative is iterative out-painting, \ie, dividing the video into temporal chunks and generating them sequentially \cite{zhong2025outdreamer}. 
This strategy enables existing short-video models to be extended to longer videos without fundamentally changing their architectures.

\begin{figure}[t]
    \centering
    \includegraphics[width=0.98\linewidth]{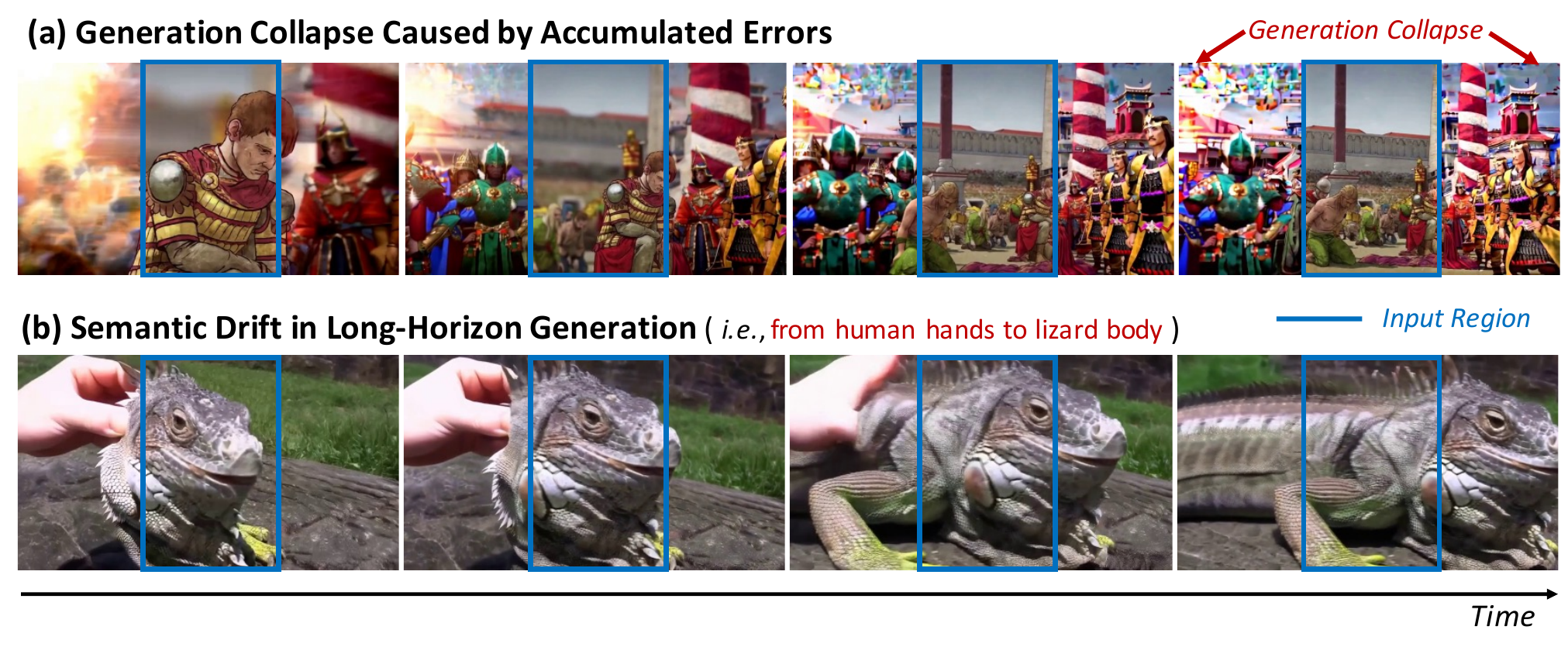}
    \vspace{-15pt}
    \caption{Illustration of the two main challenges in long video out-painting: (a) unstable generation including abrupt transitions, brightness flickering, and even complete generation collapse caused by accumulated errors, and (b) semantic drift like changed scene layouts, unstable object identities, or inconsistent subject appearances over long time. Both cases are generated by VACE model~\cite{vace}.}
    \label{fig:teaser}
    \vspace{-20pt}
\end{figure}

However, simply applying chunk-by-chunk generation introduces two major challenges over long horizons. 
First, consistency across neighboring chunks is difficult to preserve, often causing abrupt transitions, brightness flickering, and progressively accumulated errors that may eventually lead to complete generation collapse. 
As shown in \cref{fig:teaser}~(a), the out-painting on the left and right sides gradually collapses due to accumulated errors, resulting in excessive color spill and completely abstract objects. 
Second, as the temporal horizon grows longer, the generated content tends to drift semantically, resulting in changed scene layouts, unstable object identities, or inconsistent subject appearances over time. 
For example, in \cref{fig:teaser}~(b), the model initially out-painted a human hand, but the hand gradually transforms into a lizard body over a long period.

To address the above problems, we introduce \textbf{\method}, a \textbf{S}emantically \textbf{C}onsistent long video \textbf{O}ut-\textbf{P}ainting method, tackling long video out-painting from two aspects: stable long-horizon generation and semantic consistency preservation.

One fundamental requirement for long video out-painting is to maintain smooth and stable long-horizon generation, avoiding abrupt transitions, brightness flickering, and generation collapse (\ie, caused by accumulated errors). 
To this end, we propose a long-horizon stabilization strategy for iterative out-painting from three aspects. 
First, we adopt a self-forcing training scheme \cite{self_forcing}, where the model uses its own predictions as references during training, so that the training process is better aligned with inference-time iterative generation. 
Second, we introduce latent calibration by normalizing the predicted latents, which constrains their distribution from drifting too far away and thus reduces the risk of error accumulation and generation collapse. 
Third, instead of relying only on the out-painted results from the previous chunk as a reference, we further incorporate future references to provide richer temporal guidance and improve the stability of long-range generation.

Another, more advanced requirement is semantic consistency, \ie, subjects, scenes, and backgrounds should remain consistent over long temporal horizons.
Even with smooth and stable long-horizon generation, individual subjects may gradually and seamlessly transform into entirely different ones as generation proceeds, as illustrated in \cref{fig:teaser}~(b).
To address this issue, we recognize that the rich context of a long video provides a valuable source of visual references for maintaining semantic consistency.
While temporally adjacent reference frames help maintain local continuity and smooth transitions between consecutive chunks, such local references become insufficient over long temporal horizons.
They mainly provide short-range motion and appearance cues, but often fail to preserve subject identity and scene semantics across distant chunks.
We therefore build an agent based on Vision Language Models (VLMs) \cite{qwen2.5, llava, qwen3.5, cogvlm, chen2024restoreagent, zhu2025agenticir, gu2025agenticposition} that searches the entire video and automatically selects reference frames with the most informative semantic cues, providing global guidance for semantically consistent long-horizon generation.

To train and evaluate our \method, we collect a large-scale, high-quality training dataset and carefully construct a benchmark specifically for long-video out-painting. 
Our training dataset covers a wide range of video genres, consisting of 62K video clips, and ensures high-quality and diverse content that is essential for robust training across complex, dynamic scenes. 
The \bench benchmark, curated by human experts, includes representative video clips selected for their complexity and diversity, providing a comprehensive evaluation platform for testing the model's ability to maintain semantic consistency and temporal coherence in challenging long-video generation scenarios.

Extensive experimental results show the performance advantages of our \method. 
Specifically, in the \bench benchmark, we improve the FVD~\cite{fvd} by 28.19 points and PSNR by 0.92 dB compared to the previous best model. 
The ablation experiments show the effectiveness of our proposed two main modules and the design choice of our key reference frame retrieval method.

\vspace{-5pt}
\section{Related Works}
\vspace{-3pt}

\textbf{Video Generation Models.} 
Early video diffusion models extend image diffusion architectures to the temporal domain by introducing temporal attention \cite{wu2023tune}, 3D convolutions, or latent video representations, enabling the synthesis of visually realistic and temporally coherent short clips~\cite{ho2022video, blattmann2023align, zhou2022magicvideo}. Subsequent works improve generation quality and controllability by scaling training data \cite{chen2024videocrafter2}, model capacity, and conditioning signals, leading to increasingly capable text-to-video and image-to-video systems~\cite{ho2022imagen, singer2022make, animatediff}. Recently, open large-scale video generation models such as Open-Sora Plan~\cite{lin2024open} and Wan~\cite{wan2025wan} demonstrate strong generalization across diverse video creation tasks, and provide powerful foundation models for downstream video editing and completion.
In addition, to overcome the challenges of long-video generation and reduce the gap between training and inference, recent methods have explored auto-regressive generation, such as Diffusion Forcing~\cite{chen2024diffusion}, Self Forcing~\cite{self_forcing}, and Rolling Forcing \cite{liu2025rolling}. These methods have been shown to be effective in improving long-horizon video generation.

\textbf{Video Out-painting} Video out-painting seeks to extend the spatial contents of a video beyond its initial boundaries. Existing methods usually focus on temporal consistency or stronger conditioning. Complete and Temporally Consistent Video Out-painting builds background estimates, uses optical flow for temporal consistency, and introduces image shifting to improve frame completion \cite{dehan2022videooutpainting}. M3DDM is particularly relevant to our setting: it uses multiple guide frames to connect clip-level inference and extracts global frames of the video as prompts via cross-attention \cite{fan2023m3ddm}. More recent out-painting methods strengthen guidance in other ways. MOTIA performs input-specific adaptation to learn source-video patterns \cite{wang2024beyouroutpainter}. Infinite-Canvas injects the source video and its relative positional relations into each spatial window \cite{chen2025infinitecanvas}; and Unboxed combines 3D-supported out-painting for static regions with dynamic object completion and guided video synthesis \cite{yu2025unboxed}. These methods show that stronger guidance beyond the current frame is beneficial, but they still do not formulate reference selection as a query-dependent retrieval problem over the full video timeline. Moreover, VACE \cite{vace} as a general video editing method could achieve video out-painting through specific mask design.

\textbf{Reference-driven Video Completion} Recent work has also explored richer reference signals for video completion \cite{zi2025cococo}. HomoGen~\cite{ding2025homogen} propagates contextual pixels via homography and injects them as priors into a video diffusion model. 
VideoRepainter~\cite{guo2025keyframe} uses a user-edited key frame as an explicit reference and propagates its semantics to the remaining frames. 
These methods confirm the value of strong external guidance, but the guidance is either geometrically propagated or manually specified, rather than automatically retrieved from the source video according to the semantic needs. 

\vspace{-4pt}
\section{Methods}\label{sec:methods}
\vspace{-4pt}

\subsection{Preliminaries}
\vspace{-2pt}

\textbf{VACE Model.}
VACE~\cite{vace} is a general video editing model with unified support for various video editing tasks. 
The inputs of VACE are termed as video condition units, which could be expressed as:
\begin{equation}
    I = \{T, V_{in}, M\},
\end{equation}
where $T$ is the textual prompt, $V_{in}$ is the video sequence $V_{in} = \{v_0, v_1, ..., v_{n-1}\}$ with each $v_i \in \mathbb{R}^{H \times W \times 3}$, and $M$ is the mask sequence $M = \{m_0, m_1, ..., m_{n-1}\}$ with each $m_i \in \{0, 1\}^{H \times W}$. 
In $M$, $0$ and $1$ indicate the pixels that should be maintained and edited, respectively. 
The VACE model $f$ takes $I = \{T, V_{in}, M\}$ as inputs and then outputs an edited video $V_{out}$ as: 
\begin{equation}
    V_{out} = f(I = \{T, V_{in}, M\}).
\end{equation}
VACE is trained on extensive and large-scale datasets, thus could generalize to different masks very well.
By designing different masks, VACE could edit regions at various scales and positions.

\textbf{VACE for Video Out-painting} 
We could design inputs $I = \{T, V_{in}, M\}$ to achieve video out-painting using the VACE model. 
For input video $V_{in}$, we zero-pad the original video to the out-painting target resolution, thus the input video $V_{in}$ will have black borders as in \cref{fig:selfforcing}. 
For masks $M$, we assign the mask values of the original region as 0, meaning these pixels should be kept. 
The mask values of the out-painted region (\ie, the black border) are assigned as 1, indicating that they should be generated.  
The textual prompts $T$ are obtained by VLM captioning on the original video. 
With above designed inputs, VACE could serve as a baseline model for video out-painting.

\textbf{VACE Out-painting with Reference Frames} 
If we have an out-painted frame with the target resolution, an intuitive idea is to treat the out-painted frame as the reference frame to guide a smooth and continuous editing. 
We could achieve this simply by appending an out-painted frame before the input video $V_{in}$ and an all 0 mask before the masks $M$, since the all 0 mask means that all pixels should be kept and unchanged. 
Multiple reference frames could be added in the same way.

\textbf{VACE Chunk-by-Chunk Out-painting} 
With the above preliminaries, we could implement a chunk-by-chunk video out-painting pipeline. 
For a long input video, we first divide it into several consecutive chunks and apply VACE out-painting to each chunk in temporal order. 
The generated results from previous chunks are then used as reference frames for subsequent chunks, forming a simple chunk-by-chunk baseline for long video out-painting.

\vspace{-5pt}
\subsection{Foundation: Smooth and Stable Long-Horizon Generation}
\vspace{-5pt}

A fundamental requirement for long video out-painting is to ensure smooth and stable long-horizon generation, avoiding abrupt transitions, brightness flickering, and generation collapse caused by accumulated errors.
However, as the generation proceeds over time, the out-painted regions gradually degrade due to error accumulation, leading to severe color spillover and distorted structures (See \cref{fig:teaser}(a)).
To address this, we propose a long-horizon stabilization strategy for iterative out-painting.

\begin{wrapfigure}{r}{0.6\textwidth}
    \vspace{-12pt}
    \centering
    \includegraphics[width=1.0\linewidth]{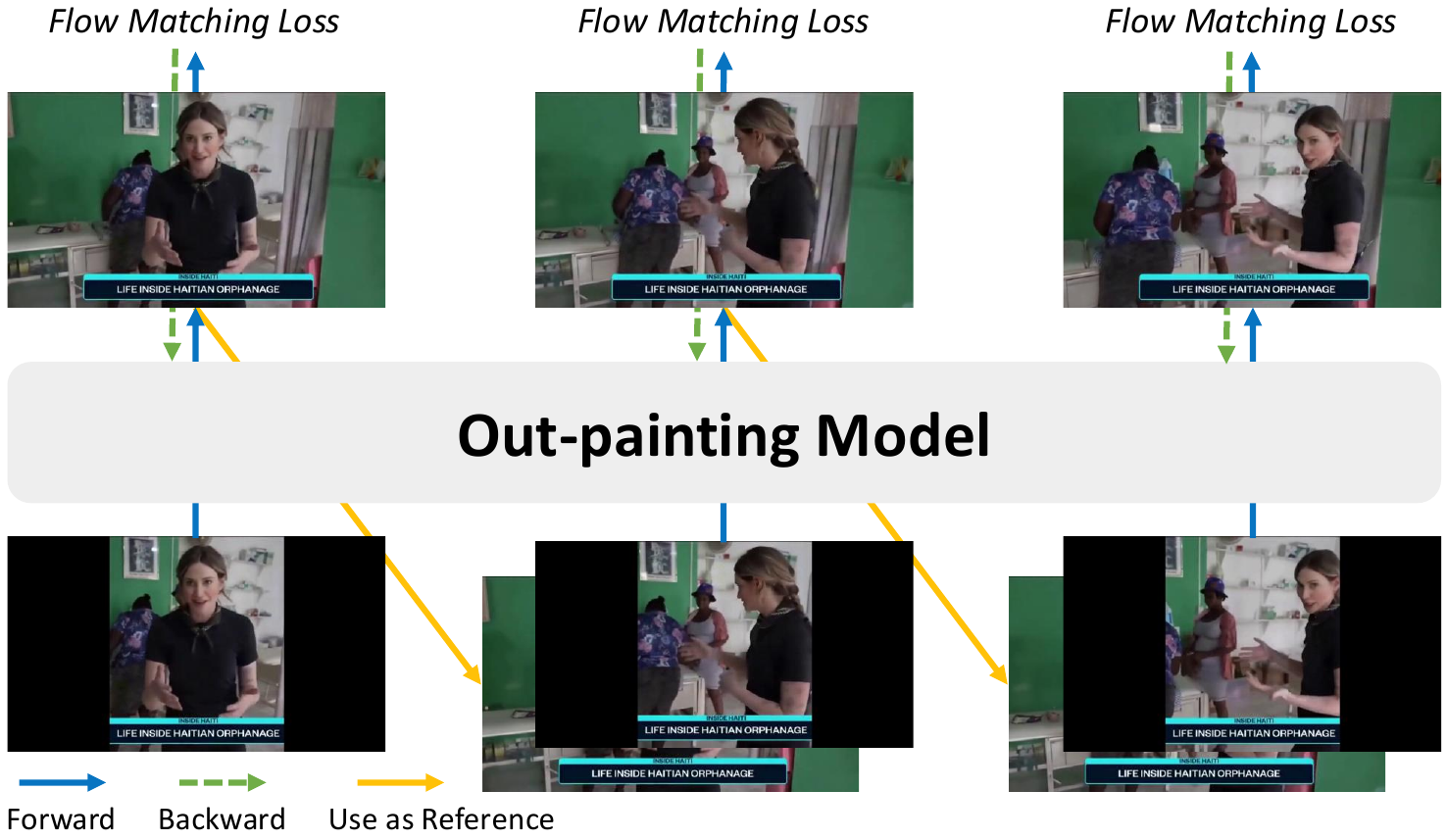}
    \caption{Illustration of self-forcing training. During training, we use the model-predicted frames as reference frames, which could reduce the gap between training and inference.}
    \label{fig:selfforcing}
    \vspace{-10pt}
\end{wrapfigure}
\textbf{Self-forcing Training with Model-predicted References} 
Chunk-by-chunk video out-painting requires auto-regressive generation across multiple chunks, making training-inference mismatch a major source of error accumulation. During training, if the model always conditions on ground-truth history frames, it only observes clean reference inputs. At inference time, however, later chunks depend on previously generated outputs, which inevitably contain noise and bias. This discrepancy grows over time and leads to unstable long-horizon generation.

To reduce this gap, inspired by~\cite{self_forcing}, as depicted in \cref{fig:selfforcing}, we introduce a self-forcing training strategy. During training, we construct two consecutive chunks. The model first performs online out-painting on the first chunk and produces predicted latent representations. These latents are then decoded into pixel space through the VAE decoder, and the last several generated frames are pre-pended to the second chunk as prefix conditions. Importantly, these prefix frames are not out-painted again; instead, they serve only as temporal conditioning signals. By exposing the model to its own predictions during training, self-forcing better aligns training with auto-regressive inference and improves long-range stability.

\textbf{Latent Calibration} 
A second source of error accumulation comes from the latent distribution shift over time. 
In long-video chunk-wise inference, generated frames must be repeatedly written back and re-encoded into latent space. Such repeated operations can progressively distort latent statistics, causing degradation in brightness, saturation, contrast, and texture fidelity.

To address this issue, we introduce latent calibration. We treat the latent statistics of several real frames from the first chunk as an anchor distribution and compute their mean $\mu_{real}$ and variance $\sigma^2_{real}$. 
During the out-painting process of next chunk, we align the reference latent $z_{raw}$ to the anchor distribution to obtain the final reference latent $z$ as: 
\begin{equation}
    \hat{z} =
    \left(
    \frac{z_{\mathrm{raw}} - \mu_{\mathrm{raw}}}
    {\sigma_{\mathrm{raw}} + \epsilon}
    \right)
    \sigma_{\mathrm{real}}
    + \mu_{\mathrm{real}}, 
    \qquad
    z = \alpha \hat{z} + (1 - \alpha) z_{\mathrm{raw}}, 
\end{equation}
where $\mu_{\mathrm{raw}}$ and $\sigma_{\mathrm{raw}}$ are the mean and standard deviation of the raw reference latent $z_{\mathrm{raw}}$, respectively, and $\epsilon$ is a small constant $10^{-6}$ to prevent numerical instability caused by division by zero, $\alpha$ is a weighted hyper-parameter and is set as 0.9. 
In this way, the latent space remains close to the original real-video statistics throughout long-horizon generation, which reduces long-term quality degradation and lowers the risk of generation collapse.
Note that the latent calibration is conducted during both self-forcing training and inference process.

\textbf{Bidirectional Reference} 
In standard chunk-wise inference, the current chunk usually conditions only on past frames. Although such one-sided conditioning helps maintain local temporal smoothness, it does not provide any information about future content. As a result, the content synthesized in the current chunk may become inconsistent with future chunks, leading to flickering and semantic discontinuities near chunk boundaries.

We therefore introduce bidirectional reference chunk inference. 
When out-painting the current chunk, we not only condition on historical frames but also include a small set of future frames as auxiliary references. 
The future reference could be implemented by a quite simple inference trick \textbf{without the need of retraining}. 
Specifically, during inference, we increase the input video length to cover a longer time than the current chunk. 
After out-painting, we keep only the outputs corresponding to the current chunk and discard the results of future frames. 
Although simple, this strategy allows the model to access a broader temporal context and anticipate scene evolution during generation, effectively reducing boundary inconsistency.

\vspace{-5pt}
\subsection{Advance: Semantic Consistency via Key Reference Frame Retrieval}
\vspace{-5pt}

Even with smooth and stable long-horizon generation, some subjects may gradually drift into totally different ones as generation proceeds, as shown in \cref{fig:teaser}(b). 
Therefore, a more advanced requirement is semantic consistency, \ie, the subjects, scenes, and backgrounds should remain consistent over long temporal horizons. 
Here we show that good reference frames could provide semantically consistent information, and design a method to automatically select the most suitable reference frames.

\begin{figure}[t]
    \centering
    \includegraphics[width=1.0\linewidth]{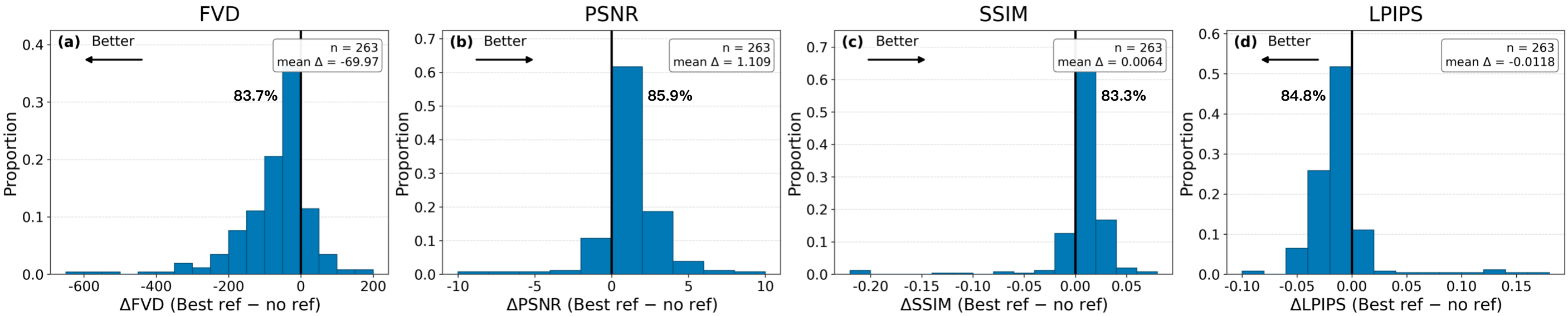}
    \vspace{-15pt}
    \caption{
    Illustration of the importance of suitable reference frame. 
    Adding only one reference frame could significantly improve the out-painting performance of the whole video by nearly 70 FVD points and more than 1.1 dB PSNR on average. 
    }
    \label{fig:hist}
    \vspace{-15pt}
\end{figure}

\textbf{What a Good Reference Can Do}
We restrict reference frames to real frames sampled from the input video, rather than using externally generated images. This design has two advantages. First, it avoids introducing synthetic artifacts or additional distribution mismatch. Second, these frames naturally share the same subjects, scene layout, and visual style as the target video, making them more suitable as conditioning signals for out-painting. For long videos, relying only on temporally adjacent history frames provides local continuity but is often insufficient for preserving global semantics over long horizons. In contrast, well-chosen reference frames can provide complementary structural cues, appearance information, and contextual semantics, thereby mitigating facial distortion, background inconsistency, and semantic drift across chunks.

We demonstrate the effectiveness of reference frames on \bench benchmark by comparing each sample with and without reference frames across multiple metrics in \cref{fig:hist}. 
Here the reference frames are sampled from the input video at one frame per second. 
We report the gains obtained by the best reference frames of each sample. 
As shown in \cref{fig:hist}, most samples (around 83\% to 86\%) achieve better results with reference frames than without them across all metrics. 
In particular, adding only one reference frame could significantly improve the out-painting performance of the whole video by nearly 70 FVD and more than 1.1 dB PSNR on average, indicating a substantial gain in video quality.

\textbf{What Makes a Good Reference Frame}
Our analysis suggests that a good reference frame should satisfy two key properties: \textbf{appropriate spatial overlap} and \textbf{information usefulness}.
Spatial overlap determines whether the candidate frame can be reliably aligned with the query chunk, while information usefulness indicates whether the candidate provides content that is truly helpful for recovering the missing out-painting region.
These two properties are complementary: the former ensures usable geometric correspondence, and the latter ensures effective contextual guidance.

Spatial overlap reflects the degree of shared visual content between the candidate reference frame and the query chunk.
If the overlap is too large, the candidate frame mostly repeats the already visible region and contributes limited new information for out-painting.
If the overlap is too small, the candidate may become difficult to align with the query chunk, making its structural guidance unreliable or even misleading.
Therefore, an effective reference frame should maintain a moderate overlap, balancing reliable spatial correspondence with sufficient complementary context.

Information usefulness measures whether the complementary content provided by the reference frame is actually beneficial for the missing region.
Even when a candidate frame has an appropriate spatial overlap, it may still be less useful if its newly introduced content is irrelevant to the area to be out-painted.
A good reference should not only introduce additional content, but also provide the right contextual cues needed for out-painting.
Thus, the most valuable reference frames are those that are both spatially compatible with the query chunk and informative for the target missing region.

\begin{figure}[t]
    \centering
    \includegraphics[width=1.0\linewidth]{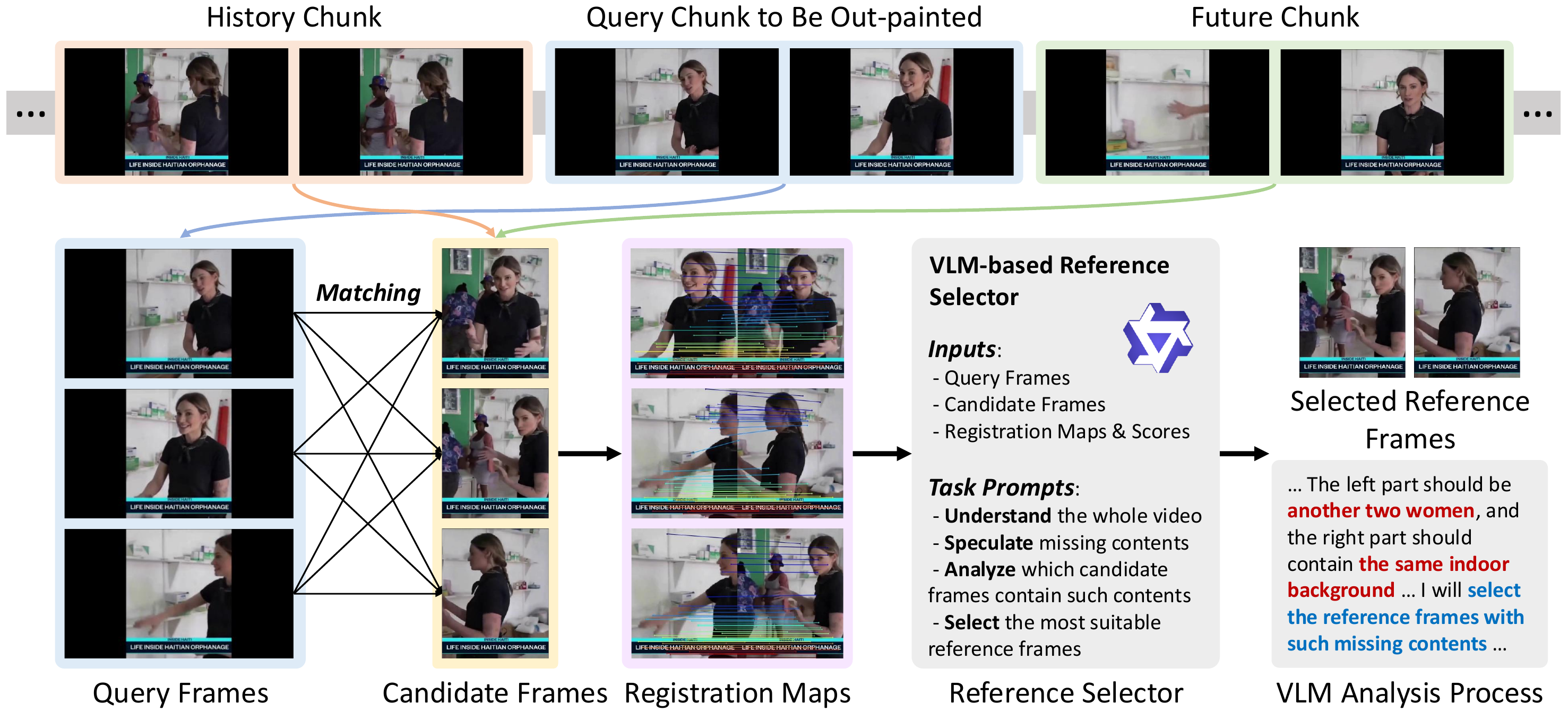}
    \vspace{-15pt}
    \caption{
    Illustration of key reference frame retrieval framework. 
    Given query frames from current query chunk to be out-painted and candidate frames from both history chunk and future chunk, we aim to find the most suitable reference frames to out-paint current chunk. 
    Specifically, we calculate the registration maps and scores for all query-candidate pairs, and then input them into a VLM, Qwen3.5-VL-27B~\cite{qwen3.5}. 
    The VLM is requested to understand the whole video, speculate the missing contents (\ie, contents to out-paint like the two women and the indoor background), analyze which candidate frames may contain such missing contents, finally select the most suitable reference frames. 
    }
    \label{fig:reference}
    \vspace{-15pt}
\end{figure}

\textbf{How to Get a Good Reference Frame}
Based on the above observations, we introduce a geometry-guided and semantics-aware reference retrieval method.
For each query chunk, we uniformly sample four query frames to capture its missing-region context, and then extract candidate frames from the full video at fixed temporal intervals and rank them through a two-stage procedure.

In the first stage, we focus on the \textbf{appropriate spatial overlap}. 
As depicted in \cref{fig:reference}, we first use the image registration model Mast3R \cite{mast3r_eccv24} to estimate the spatial relationship between each query frame and candidate frame. 
Mast3R produces registration scores that reflect the degree of geometric match. 
We use these scores to remove poorly aligned candidates (\ie, bottom 30\% of poorly aligned candidates) before semantic reasoning. 
This geometric filtering stage substantially reduces the search space and ensures that the remaining candidates satisfy a basic level of spatial compatibility.

In the second stage, the main target is \textbf{information usefulness}. 
Since information usefulness is hard to measure by a simple metric, we resort to a VLM, Qwen3.5-VL-27B~\cite{qwen3.5}. 
Unlike registration scores, which mainly reflect geometric similarity, the VLM can reason about which visual cues are missing in the out-painting region and whether a candidate frame can provide those cues \cite{alayrac2022flamingo, ye2023mplug, you2024depictqa, you2025edqa, you2026photoframer}. 
Specifically, as shown in \cref{fig:reference}, the VLM jointly examines the query frames, candidate frames, and registration maps and scores to determine whether a candidate can supply informative appearance, structural, or contextual content for out-painting. 
This cascaded design, geometry-guided filtering followed by semantics-aware reasoning, improves both retrieval efficiency and selection accuracy. 
The selected key reference frame is then fed into the model as an additional condition for generation.

\textbf{Joint Training with Selected Key Reference Frame}
Rather than using the selected key reference frames only as a test-time plug-in, we incorporate them into the training stage to make the out-painting model better adapted to reference-guided generation.
Specifically, we sample a subset of training videos and construct training chunks with their retrieved key reference frames, following the same reference-format design used in inference.
The selected frame is appended to the input video as an additional reference condition with an all-zero mask, so that the model learns to exploit its visual cues while keeping the reference content unchanged.
This joint training strategy exposes the model to the same type of selected reference conditions during optimization and inference, reducing the mismatch between training and inference.

\vspace{-5pt}
\section{Experiments}\label{sec:exp}
\vspace{-5pt}

\subsection{Dataset and Benchmark Construction}
\vspace{-3pt}

\textbf{Training Dataset Collection} 
(a) \textit{Source video collection}. 
To construct a high-quality training dataset, we collect a large amount of high-resolution videos covering diverse categories, including portraits, architecture, animals, plants, and natural scenes. 
After removing videos with black borders, low effective resolution, insufficient bitrate, or visible compression artifacts, we segment videos into scenes using PySceneDetect~\cite{pyscenedetect} and retain scenes longer than 64 frames at 24 FPS. 
We then evaluate each clip by frame quality and scene dynamics, where frame quality is measured by averaged MUSIQ~\cite{musiq} and LAION-Aesthetics~\cite{laion_aesthetics} scores from frames sampled at 3 FPS, and scene dynamics are measured by RAFT optical-flow magnitude~\cite{raft}. 
By jointly ranking these criteria, we select 62K high-quality clips. 
All selected clips are down-sampled by a factor of two and extracted at 24 FPS as individual frames for subsequent training.
(b) \textit{Video captioning}. 
For each clip, we further obtain a textual caption using VLM captioning~\cite{qwen3.5}, which is used as the textual prompt for training. 
(c) \textit{Random mask for video out-painting}. 
Based on these high-quality videos, we construct video out-painting training samples by applying random exterior masks. 
Specifically, we randomly apply the random mask strategy using a mask ratio between 0.25 and 0.66. 
The unmasked region is always constrained to be a complete rectangle, corresponding to the original visible video content, while the masked region only appears outside this rectangle. 
This design matches the practical setting of video out-painting and allows the model to generalize to diverse out-painting ratios and spatial layouts.

\textbf{Existing Benchmarks} 
To verify the effectiveness of \method, we conduct evaluations on DAVIS~\cite{pont20172017} and YouTube-VOS~\cite{xu2018youtube}, which are widely used benchmarks for video out-painting. Following M3DDM~\cite{fan2023m3ddm}, we compare the results of different methods in the horizontal direction, using mask ratios of 0.25 and 0.66.

\textbf{\bench Benchmark Construction} 
We construct our \bench benchmark (short for \textbf{L}ong \textbf{V}ideo \textbf{O}ut-painting) from existing LongVideoBench \cite{wu2024longvideobench} and Video-MME \cite{fu2025mme} benchmarks using a multi-stage filtering pipeline designed specifically for reference-based long-video out-painting. We first apply scene boundary detection to split each source video into temporally coherent shots, and retain only sufficiently long segments so that temporally distant reference retrieval is meaningful. We then remove degenerate cases with little visual change over time, such as news-style or talking-head footage, by sampling frames, uniformly within each segment and measuring content variation together with motion statistics. Segments with insufficient temporal change are discarded, as are clips with overly violent motion that would make the task ill-posed.

To ensure suitability for left-right out-painting, we further apply geometric validation. We estimate inter-frame motion using optical flow and robust point tracking, and retain clips with stable horizontal camera motion, limited rotation, and small scale drift. We then verify whether temporally distant frames can align with target frames and provide sufficient recoverable content near the left or right borders. Clips with unreliable geometry or weak recoverable regions are discarded. As a result, LVO-Bench focuses on long, dynamic, and geometrically valid videos where reference-frame retrieval is both necessary and useful.

\textbf{Benchmark Composition and Scene Diversity}
The final \bench contains 52 video cases, with durations ranging from approximately 5 seconds to more than 20 seconds.
We examine its diversity along two complementary dimensions: scene environment and camera shot scale.
The benchmark contains 26 indoor and 26 outdoor videos.
It also includes 3 close-up, 36 medium-shot, and 13 long-shot videos.
These counts describe two categorizations of the same 52 cases.
The combination of different environments, framing scales, and temporal lengths provides varied conditions for evaluating long-range reference retrieval and video out-painting.

\vspace{-5pt}
\subsection{Metrics and Implementation Details}
\vspace{-2pt}

\textbf{Metrics} Our evaluation approach utilizes four well-established metrics: Peak Signal to Noise Ratio (PSNR), Structural Similarity Index Measure (SSIM)~\cite{ssim}, Learned Perceptual Image Patch Similarity (LPIPS)~\cite{lpips}, and Frechet Video Distance (FVD)~\cite{fvd}. For assessing PSNR, SSIM, and FVD, the generated videos are converted into frames within a normalized value range of [0, 1]. LPIPS is evaluated over a range of [-1, 1]. About the FVD metric, we adopt a uniform frame sampling, with 16 frames per video for evaluation following M3DDM \cite{fan2023m3ddm, gu2020pipal}.

\textbf{Implementation Details} 
Our model is built upon and initialized from VACE-1.3B~\cite{vace}. The LoRA~\cite{lora} technique is adopted. The LoRA rank $r$ and LoRA $\alpha$ are set to 16 and 8, respectively. The low-rank adapters are trained using the Adam optimizer~\cite{adam}. We set the learning rate to 0.0001, and set the weight decay to 0.01. The model is trained on 8 NVIDIA A6000 GPUs for 20 hours.

\vspace{-7pt}
\subsection{Results on Benchmarks}
\vspace{-3pt}

\textbf{Quantitative Results on DAVIS and YouTube-VOS Benchmarks}
In~\cref{tab:res_davis_youtube}, we evaluate the video out-painting performance of our \method on the DAVIS~\cite{pont20172017} and YouTube-VOS~\cite{xu2018youtube} benchmarks. 
Note that the two benchmarks mainly contain \textit{short clips} where \textbf{most samples only have one chunk} (\ie, $\leq$ 3 seconds for DAVIS and 3 to 5 seconds for YouTube-VOS).
Although LongSCOP is specifically designed for long video out-painting with multiple chunks, it still achieves competitive results. 
Compared to the base VACE, the performance is improved stably in SSIM, PSNR, and LPIPS. 
For video-level distribution similarity, LongSCOP achieves a comparable FVD to VACE, and clearly better than other task-specific video out-painting methods.

\begin{table}[t]
    \centering
    \scriptsize
    \setlength{\tabcolsep}{8.2pt}
    \caption{
    Results of video out-painting performance on the DAVIS~\cite{pont20172017} and YouTube-VOS~\cite{xu2018youtube} benchmarks across multiple metrics. The best and second best results are \textbf{bold} and \underline{underlined}. Note that the two benchmarks mainly contain \textit{short clips} where \textbf{most samples only have one chunk} (\ie, $\leq$ 3 seconds for DAVIS and 3 to 5 seconds for YouTube-VOS). Though specifically designed for long videos with multiple chunks, we still achieve comparable results on such short clips.
    }
    \label{tab:res_davis_youtube}
    \vspace{-5pt}
    \begin{tabular}{l|cccc|cccc}
        \toprule
        & \multicolumn{4}{c|}{DAVIS~\cite{pont20172017}} & \multicolumn{4}{c}{YouTube-VOS~\cite{xu2018youtube}} \\
        \cmidrule(lr){2-5} \cmidrule(l){6-9}
        & SSIM $\uparrow$ & PSNR $\uparrow$ & LPIPS $\downarrow$ & FVD $\downarrow$
        & SSIM $\uparrow$ & PSNR $\uparrow$ & LPIPS $\downarrow$ & FVD $\downarrow$ \\
        \midrule
        Dehan \cite{dehan2022videooutpainting}      & 0.627 & 17.96 & 0.233 & 363.1 & 0.720 & 18.25 & 0.228 & 149.7 \\
        SDM \cite{rombach2022high}        & 0.708 & 20.02 & 0.217 & 334.6 & 0.728 & 19.91 & 0.200 & 94.8 \\
        M3DDM \cite{fan2023m3ddm}     & 0.708 & 20.26 & 0.203 & 300.0 & 0.731 & \underline{20.20} & \underline{0.185} & 66.6 \\
        Infinite-Canvas \cite{chen2025infinitecanvas} & 0.726 & \textbf{20.80} & \textbf{0.160} & 242.8 & - & - & - & - \\
        OutDreamer \cite{zhong2025outdreamer} & \textbf{0.757} & \underline{20.30} & 0.174 & 268.9 & \textbf{0.764} & \textbf{20.21} & \textbf{0.183} & 56.0 \\
        VACE \cite{vace} & 0.696 & 18.83 & 0.257 & \textbf{195.7}
                   & 0.739 & 18.97 & 0.234 & \textbf{45.8} \\
        \midrule
        LongSCOP (Ours) & \underline{0.754} & 20.04 & \textbf{0.160} & \underline{197.1}
        & \underline{0.748} & 19.31 & 0.231 & \underline{53.9} \\
        \bottomrule
    \end{tabular}
    \vspace{-4pt}
\end{table}

\begin{figure}[t]
    \centering
    \includegraphics[width=1.0\linewidth]{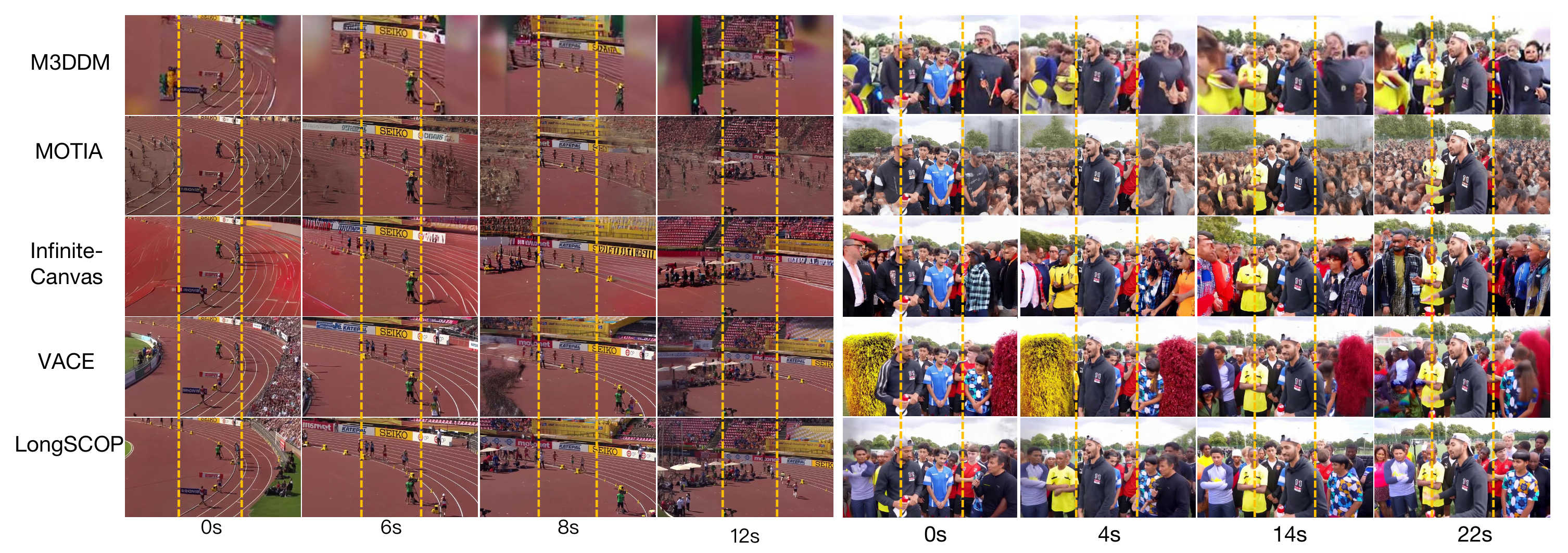}
    \vspace{-20pt}
    \caption{
    Qualitative comparison. 
    Our \method could achieve stable video out-painting results and maintain the semantic consistency over a long time horizon. The region circled by the yellow box indicates the original input region. Zoom in for better views. 
    }
    \label{fig:res_compare}
    \vspace{-5pt}
\end{figure}

\textbf{Quantitative Results on our \bench Benchmark}
The video out-painting results on our \bench benchmark are illustrated in \cref{tab:res_lvo}. 
Our \method improves the FVD by 28.19 points (197.28 \vs 225.47) and PSNR by 0.92 dB (19.87 \vs 18.95) compared to the previous best model. 
These clear advantages prove the superiority of our \method.

\textbf{Results across Scene Categories.}
We further evaluate \method across scene categories in \bench, beyond the aggregate results in \cref{tab:res_lvo}.
The benchmark includes 26 indoor and 26 outdoor videos, as well as 3 close-up, 36 medium-shot, and 13 long-shot videos.
As shown in \cref{tab:lvo_scene_fvd}, LongSCOP improves FVD in every category.
The outdoor subset improves from 306.50 to 236.64.
Close-up videos show a larger numerical gain, but this subset contains only three cases.

\begin{table}[!t]
    \centering
    \begin{minipage}[t]{0.51\textwidth}
        \centering
        \caption{Video out-painting results on \bench.}
        \label{tab:res_lvo}
        \vspace{-5pt}
        \scriptsize
        \setlength{\tabcolsep}{4.1pt}
        \begin{tabular}{ c|cccc}
            \toprule
            Method & SSIM$\uparrow $ & PSNR$\uparrow$ & LPIPS$\downarrow$ & FVD$\downarrow$  \\ 
            \midrule
            M3DDM           & 0.505 & 17.16 & 0.220  & 270.3 \\
            MOTIA           & 0.647 & 16.64 & 0.320  & 419.4 \\
            Infinite-Canvas & 0.731 & 18.95 & 0.237  & 299.3 \\
            VACE            & 0.725 & 18.59 & 0.238  & 225.5 \\
            \bottomrule
            LongSCOP & \textbf{0.749} & \textbf{19.87} & \textbf{ 0.207} & \textbf{197.3} \\ \bottomrule
            \end{tabular}
    \end{minipage}
    \hfill
    \begin{minipage}[t]{0.46\textwidth}
        \centering
        \caption{Scene-wise FVD on \bench.}
        \label{tab:lvo_scene_fvd}
        \vspace{-5pt}
        \scriptsize
        \setlength{\tabcolsep}{2.7pt}
        \renewcommand{\arraystretch}{1.05}
        \begin{tabular}{@{}lrrr@{}}
            \toprule
            Category & $N$ & VACE & LongSCOP \\
            \midrule
            Indoor      & 26 & 306.17 & \textbf{299.65} \\
            Outdoor     & 26 & 306.50 & \textbf{236.64} \\
            Close-up    & 3  & 421.53 & \textbf{296.03} \\
            Medium-shot & 36 & 256.12 & \textbf{217.00} \\
            Long-shot   & 13 & 445.53 & \textbf{401.26} \\
            \bottomrule
        \end{tabular}
    \end{minipage}
    \vspace{-5pt}
\end{table}

\textbf{Qualitative Results}
As shown in \cref{fig:res_compare}, our method is significantly superior to other methods in terms of both stability and semantic consistency. 
First, our \method keeps a stable generation over the whole video, while all baseline methods suffer from a degree of generation collapse. 
Moreover, in \cref{fig:res_compare}(right), \method maintains almost the same characteristics of the man in yellow and the man in purple throughout all 22 seconds, while all baselines fail to maintain such semantic consistency.

\vspace{-4pt}
\subsection{Ablation Studies}
\vspace{-3pt}

\textbf{Ablation of Our Proposed Two Main Modules}
We ablate the two main modules, including stable generation strategies and reference frame retrieval method, in \cref{tab:ablation_main}. 
The stable generation strategies improve SSIM from 0.725 to 0.751, PSNR from 18.59 to 19.47, and FVD from 225.5 to 208.6, confirming their effectiveness in reducing long-horizon degradation. 
Adding reference frame retrieval further improves PSNR to 19.87 and FVD to 197.3, demonstrating better semantic consistency.

\vspace{-5pt}

\begin{table}[!tbp]
    \centering
    \begin{minipage}[c]{0.28\linewidth}
        \caption{Ablation of stable generation and reference retrieval on \bench.}
        \label{tab:ablation_main}
    \end{minipage}\hfill
    \begin{minipage}[c]{0.69\linewidth}
        \centering
        \scriptsize
        \setlength{\tabcolsep}{6.2pt}
        \resizebox{\linewidth}{!}{
            \begin{tabular}{cc|cccc}
                \toprule
                Stable Gen. & Refer. Retrieval & SSIM$\uparrow$ & PSNR$\uparrow$ & LPIPS$\downarrow$ & FVD$\downarrow$ \\
                \midrule
                \ding{55} & \ding{55} & 0.725 & 18.59 & 0.238 & 225.5 \\
                \ding{51} & \ding{55} & \textbf{0.751} & 19.47 & 0.208 & 208.6 \\
                \ding{51} & \ding{51} & 0.749 & \textbf{19.87} & \textbf{0.207} & \textbf{197.3} \\
                \bottomrule
            \end{tabular}
        }
    \end{minipage}
    \vspace{-14pt}
\end{table}

\begin{table}[t]
    \centering
    \scriptsize
    \setlength{\tabcolsep}{8.4pt}
    \caption{
    Quantitative results of various implementation variants of reference frame retrieval methods.
    }
    \label{tab:ablation_reference}
    \vspace{-5pt}
    \begin{tabular}{l|ccccccc}
    \toprule
    Method & Hit@1$\uparrow $ & Hit@3 $\uparrow$ & Recall@3 $\uparrow$ & Recall@5 $\uparrow$ & GT rank@1 $\downarrow$ & GT rank@3 $\downarrow$ & nDCG@3 $\uparrow$\\
    \midrule
    Variant (1) & 0.1394 & 0.6653 & 0.3320 & 0.5252 & 6.7849 & 7.2363 & 0.6066\\
    Variant (2) & 0.1163 & 0.5814 & 0.3036 & 0.4774 & 7.5349 & 7.4496 & 0.5612\\
    Variant (3) & 0.1395 & 0.6162 & 0.3023 & 0.4942 & 7.0076 & 7.4199 & 0.5815\\
    Variant (4) & 0.1240 & 0.5890 & 0.3056 & 0.4941 & 6.8992 & 7.4670 & 0.5590\\
    Variant (5) & 0.1434 & 0.6550 & 0.3198 & 0.5110 & 6.9729 & 7.4477 & \textbf{0.5931}\\
    Variant (6) & 0.1589 & 0.6279 & 0.3114 & 0.4994 & 6.9884 & 7.4625 & 0.5768\\
    Variant (7) & 0.1473 & \textbf{0.6744} & \textbf{0.3256} & \textbf{0.5265} & \textbf{6.5969} & \textbf{7.0853} & \textbf{0.6115}\\
    \bottomrule
    \end{tabular}
    \vspace{-10pt}
\end{table}

\begin{table}[!t]
    \centering
    \begin{minipage}[t]{0.46\textwidth}
        \centering
        \caption{Retrieval generalization across registration models and Qwen3.5 VLMs. 9B and 27B represent the number of parameters used in the Qwen model.}
        \label{tab:backbone_generalization}
        \vspace{-5pt}
        \scriptsize
        \setlength{\tabcolsep}{4.2pt}
        \renewcommand{\arraystretch}{1.06}
        \begin{tabular}{@{}lccc@{}}
            \toprule
            Registration + VLM & Hit@3$\uparrow$ & Rec@5$\uparrow$ & nDCG@3$\uparrow$ \\
            \midrule
            Mast3R + 9B  & 0.6202 & 0.4929 & 0.5890 \\
            RoMa + 9B    & 0.6473 & 0.5136 & \textbf{0.6127} \\
            RoMa + 27B   & 0.6512 & 0.4845 & 0.5965 \\
            Mast3R + 27B & \textbf{0.6744} & \textbf{0.5265} & 0.6115 \\
            \bottomrule
        \end{tabular}
    \end{minipage}
    \hfill
    \begin{minipage}[t]{0.46\textwidth}
        \centering
        \caption{Effect of registration-based candidate filtering ratios.}
        \label{tab:filtering_ratio}
        \vspace{-2pt}
        \scriptsize
        \setlength{\tabcolsep}{4.2pt}
        \renewcommand{\arraystretch}{1.04}
        \begin{tabular}{@{}lccc@{}}
            \toprule
            Removed & Hit@3$\uparrow$ & Rec@5$\uparrow$ & nDCG@3$\uparrow$ \\
            \midrule
            20\% & \textbf{0.6744} & 0.5142 & 0.5980 \\
            30\% & \textbf{0.6744} & \textbf{0.5265} & \textbf{0.6115} \\
            40\% & 0.6550 & 0.5032 & 0.5814 \\
            50\% & 0.6705 & 0.5174 & 0.5877 \\
            60\% & 0.6317 & 0.5065 & 0.5670 \\
            70\% & 0.5968 & 0.5065 & 0.5318 \\
            80\% & 0.5310 & 0.5045 & 0.4805 \\
            \bottomrule
        \end{tabular}
    \end{minipage}
    \vspace{-10pt}
\end{table}

\textbf{Ablation of Key Reference Retrieval Methods}
To systematically evaluate the design choices in the retrieval module, we construct a set of implementation variants about candidate filtering strategy, the use of registration-derived auxiliary information, and the organization of VLM reasoning. 
Specifically, we consider the following variants: (1) using manually selected reference frames as an upper-bound baseline; (2) feeding all candidate frames and the four query frames into the VLM for direct retrieval; (3) filtering out the bottom 30\% of candidates according to registration scores before passing the remaining candidates and query frames to the VLM; (4) filtering out the bottom 50\% of candidates according to registration scores before passing the remaining candidates and query frames to the VLM; (5) filtering out the bottom 30\% of candidates before VLM reasoning, and input registration scores; (6) filtering out the bottom 50\% of candidates before VLM reasoning, and input registration scores and rank information; (7) filtering out the bottom 30\% of candidates and additionally providing registration scores, rank information, and registration visualizations to the VLM.
These variants span a spectrum from purely geometry-based selection to deep integration of geometric filtering and semantic reasoning, allowing us to isolate the contribution of each design choice. 

\vspace{-5pt}

As demonstrated in \cref{tab:ablation_reference}, Variant (7) achieves the best agreement with the reference-frame rankings derived from objective generation metrics, yielding the strongest performance on HiT and Recall. We therefore adopt Variant (7) as the final design: we first filter out the bottom 30\% of candidates using registration scores, and then provide the remaining candidates, query frames, registration ranking information and registration visualizations to the VLM. This result suggests that moderate geometric filtering is effective at removing clearly invalid candidates, while explicit rank-aware registration cues help the VLM focus on references that are both spatially alignable and semantically informative.

\vspace{-8pt}

\begin{wraptable}{r}{0.5\textwidth}
    \centering
    \vspace{-8pt}
    \caption{Average inference time per 15-second video (Mast3R retrieval).}
    \label{tab:inference_cost}
    \vspace{2pt}
    \scriptsize
    \setlength{\tabcolsep}{6pt}
    \renewcommand{\arraystretch}{1.08}
    \begin{tabular}{@{}lcc@{}}
        \toprule
        Method & Time (min)$\downarrow$ & Overhead \\
        \midrule
        VACE & 13.5 & -- \\
        LongSCOP (9B) & 16.5 & +22\% \\
        LongSCOP (27B) & 20.1 & +49\% \\
        \bottomrule
    \end{tabular}
    \vspace{-5pt}
\end{wraptable}

\vspace{5pt}
\textbf{Inference time.}
For approximately 15-second videos, VACE requires 13.5 minutes on an NVIDIA A6000.
Our full pipeline with Mast3R and Qwen3.5-27B takes 20.1 minutes, incurring about 49\% overhead.
This additional cost comes from reference retrieval and bidirectional generation.
Replacing the 27B VLM with 9B reduces end-to-end latency to 16.5 minutes (22\% overhead), saving 3.6 minutes per video.
These totals are end-to-end measurements; individual stages were not separately timed.

\textbf{Generality across VLMs and registration models.}
Our reference retrieval framework separates geometric filtering from semantic reasoning.
Neither component requires out-painting-specific fine-tuning, allowing pretrained backbones to be replaced directly.
We evaluate Mast3R and RoMa \cite{edstedt2024roma} for registration, together with Qwen3.5-9B and Qwen3.5-27B for semantic reasoning.
As shown in \cref{tab:backbone_generalization}, all four combinations achieve Hit@3 above 0.62 and nDCG@3 above 0.58.
Mast3R + 27B gives the highest Hit@3 (0.6744), while RoMa + 9B remains competitive (0.6473).
These results indicate flexibility across the tested pretrained backbones.
Two real-world videos outside the benchmark further support qualitative robustness.

\textbf{Sensitivity to geometric filtering.}
We vary the fraction of registration-filtered candidates from 20\% to 80\%, keeping other components fixed.
As shown in \cref{tab:filtering_ratio}, removing 30\% achieves the best Recall@5 (0.5265) and nDCG@3 (0.6115), while tying the best Hit@3 (0.6744).
Aggressive filtering removes useful candidates and gradually reduces retrieval accuracy.

\textbf{Memory cost.}
Retrieval and generation run in separate stages without modifying the VACE DiT architecture.
The video-generation stage retains approximately the same peak memory requirement as VACE.
Image registration uses about 8\,GB, while Qwen3.5-27B requires approximately 40\,GB (distributed across two servers), compared with about 13\,GB for Qwen3.5-9B during retrieval.
Thus, the smaller VLM reduces retrieval memory cost without changing the generator.

\vspace{-8pt}
\section{Conclusions and Limitations}\label{sec:conclusion}
\vspace{-5pt}

We introduce LongSCOP, a long video out-painting framework, built upon stable generation strategies, key reference frame retrieval method, large-scale training datasets, and a long video out-painting benchmark, showing clear improvement on both generation stability and semantic coherence.

\textbf{Limitations}
Despite its effectiveness, LongSCOP still has some limitations. The VLM-based retrieval module introduces additional inference cost, and its benefit depends on the availability of informative reference frames in the input video. Performance may degrade when missing regions never appear in other frames. How to reduce the retrieval time and how to enhance the robustness of the reference frame remain important directions for future work. In addition, it is important to improve the speed of self-forcing training, which is also left as our future works.


{
\small
\bibliographystyle{abbrvnat}
\bibliography{ref}

@String(CVPR= {IEEE Conf. Comput. Vis. Pattern Recog.})

@String(ICCV= {Int. Conf. Comput. Vis.})

@String(ECCV= {Eur. Conf. Comput. Vis.})

@String(NIPS= {Adv. Neural Inform. Process. Syst.})

@String(TIP  = {IEEE Trans. Image Process.})

@String(ACMMM= {ACM Int. Conf. Multimedia})

@String(ICLR = {Int. Conf. Learn. Represent.})

@String(AAAI = {AAAI})

@String(CVPRW= {IEEE Conf. Comput. Vis. Pattern Recog. Worksh.})

@String(CVPR  = {CVPR})

@String(ICCV  = {ICCV})

@String(ECCV  = {ECCV})

@String(NIPS  = {NeurIPS})

@String(TIP   = {IEEE TIP})

@String(ACMMM = {ACM MM})

@String(ICLR  = {ICLR})

@String(CVPRW= {CVPRW})

@inproceedings{dehan2022videooutpainting,
  author    = {Lo{\"i}c Dehan and Wiebe Van Ranst and Patrick Vandewalle and Toon Goedem{\'e}},
  title     = {Complete and Temporally Consistent Video Outpainting},
  booktitle = CVPRW,
  year      = {2022},
}

@inproceedings{fan2023m3ddm,
  author    = {Fanda Fan and Chaoxu Guo and Litong Gong and Biao Wang and Tiezheng Ge and Yuning Jiang and Chunjie Luo and Jianfeng Zhan},
  title     = {Hierarchical Masked 3D Diffusion Model for Video Outpainting},
  booktitle = ACMMM,
  year      = {2023},
}

@inproceedings{wang2024beyouroutpainter,
  author    = {Fu-Yun Wang and Xiaoshi Wu and Zhaoyang Huang and Xiaoyu Shi and Dazhong Shen and Guanglu Song and Yu Liu and Hongsheng Li},
  title     = {{Be-Your-Outpainter}: Mastering Video Outpainting Through Input-Specific Adaptation},
  booktitle = ECCV,
  year      = {2024},
}

@inproceedings{chen2025infinitecanvas,
  author    = {Qihua Chen and Yue Ma and Hongfa Wang and Junkun Yuan and Wenzhe Zhao and Qi Tian and Hongmei Wang and Shaobo Min and Qifeng Chen and Wei Liu},
  title     = {{Infinite-Canvas}: Higher-Resolution Video Outpainting with Extensive Content Generation},
  booktitle = AAAI,
  year      = {2025},
}

@inproceedings{yu2025unboxed,
  author    = {Zhongrui Yu and Martina Megaro-Boldini and Robert W. Sumner and Abdelaziz Djelouah},
  title     = {Unboxed: Geometrically and Temporally Consistent Video Outpainting},
  booktitle = CVPR,
  year      = {2025},
}

@inproceedings{ding2025homogen,
  author    = {Ding Ding and Yueming Pan and Ruoyu Feng and Qi Dai and Kai Qiu and Jianmin Bao and Chong Luo and Zhenzhong Chen},
  title     = {{HomoGen}: Enhanced Video Inpainting via Homography Propagation and Diffusion},
  booktitle = CVPR,
  year      = {2025},
}

@inproceedings{guo2025keyframe,
  author    = {Yuwei Guo and Ceyuan Yang and Anyi Rao and Chenlin Meng and Omer Bar-Tal and Shuangrui Ding and Maneesh Agrawala and Dahua Lin and Bo Dai},
  title     = {Keyframe-Guided Creative Video Inpainting},
  booktitle = CVPR,
  year      = {2025},
}

@article{zhong2025outdreamer,
  title={Outdreamer: Video outpainting with a diffusion transformer},
  author={Zhong, Linhao and Li, Fan and Huang, Yi and Liu, Jianzhuang and Pei, Renjing and Song, Fenglong},
  journal={arXiv preprint arXiv:2506.22298},
  year={2025}
}

@article{lin2024open,
  title={Open-Sora Plan: Open-Source Large Video Generation Model},
  author={Lin, Bin and Ge, Yunyang and Cheng, Xinhua and Li, Zongjian and Zhu, Bin and Wang, Shaodong and He, Xianyi and Ye, Yang and Yuan, Shenghai and Chen, Liuhan and others},
  journal={arXiv preprint arXiv:2412.00131},
  year={2024}
}

@article{wan2025wan,
  title={Wan: Open and advanced large-scale video generative models},
  author={Wan, Team and Wang, Ang and Ai, Baole and Wen, Bin and Mao, Chaojie and Xie, Chen-Wei and Chen, Di and Yu, Feiwu and Zhao, Haiming and Yang, Jianxiao and others},
  journal={arXiv preprint arXiv:2503.20314},
  year={2025}
}

@inproceedings{elmoghany2025survey,
  title={A survey on long-video storytelling generation: architectures, consistency, and cinematic quality},
  author={Elmoghany, Mohamed and Rossi, Ryan and Yoon, Seunghyun and Mukherjee, Subhojyoti and Bakr, Eslam Mohamed and Mathur, Puneet and Wu, Gang and Lai, Viet Dac and Lipka, Nedim and Zhang, Ruiyi and others},
  booktitle=ICCV,
  year={2025}
}

@misc{laion_aesthetics,
    title  = {{LAION-Aesthetics}},
    author = {{LAION-Aesthetics Team}},
    year   = {2022},
    url    = {https://laion.ai/blog/laion-aesthetics}
}

@misc{qwen3.5,
    title  = {{Qwen3.5}: Towards Native Multimodal Agents},
    author = {{Qwen Team}},
    year   = {2026},
    url    = {https://qwen.ai/blog?id=qwen3.5}
}

@misc{pyscenedetect,
    title  = {{PySceneDetect Library}},
    author = {{PySceneDetect Team}},
    year   = {2014},
    url    = {https://www.scenedetect.com}
}

@article{ssim,
  title={Image quality assessment: from error visibility to structural similarity},
  author={Wang, Zhou and Bovik, Alan C and Sheikh, Hamid R and Simoncelli, Eero P},
  journal=TIP,
  year={2004},
}

@inproceedings{lpips,
  title={The unreasonable effectiveness of deep features as a perceptual metric},
  author={Zhang, Richard and Isola, Phillip and Efros, Alexei A and Shechtman, Eli and Wang, Oliver},
  booktitle=CVPR,
  year={2018}
}

@inproceedings{lora,
  title={{LoRA}: Low-rank adaptation of large language models.},
  author={Hu, Edward J and Shen, Yelong and Wallis, Phillip and Allen-Zhu, Zeyuan and Li, Yuanzhi and Wang, Shean and Wang, Liang and Chen, Weizhu and others},
  booktitle=ICLR,
  year={2022}
}

@inproceedings{vace,
  title={{VACE}: All-in-one video creation and editing},
  author={Jiang, Zeyinzi and Han, Zhen and Mao, Chaojie and Zhang, Jingfeng and Pan, Yulin and Liu, Yu},
  booktitle=ICCV,
  year={2025}
}

@inproceedings{musiq,
  title={{MUSIQ}: Multi-scale image quality transformer},
  author={Ke, Junjie and Wang, Qifei and Wang, Yilin and Milanfar, Peyman and Yang, Feng},
  booktitle=ICCV,
  year={2021}
}

@inproceedings{raft,
  title={{RAFT}: Recurrent all-pairs field transforms for optical flow},
  author={Teed, Zachary and Deng, Jia},
  booktitle=ECCV,
  year={2020}
}

@article{fvd,
  title={Towards accurate generative models of video: A new metric \& challenges},
  author={Unterthiner, Thomas and Van Steenkiste, Sjoerd and Kurach, Karol and Marinier, Raphael and Michalski, Marcin and Gelly, Sylvain},
  journal={arXiv preprint arXiv:1812.01717},
  year={2018}
}

@inproceedings{self_forcing,
  title={{Self Forcing}: Bridging the train-test gap in autoregressive video diffusion},
  author={Huang, Xun and Li, Zhengqi and He, Guande and Zhou, Mingyuan and Shechtman, Eli},
  booktitle=NIPS,
  year={2025}
}

@inproceedings{llava,
  title={Visual instruction tuning},
  author={Liu, Haotian and Li, Chunyuan and Wu, Qingyang and Lee, Yong Jae},
  booktitle=NIPS,
  year={2024}
}

@inproceedings{cogvlm,
  title={{CogVLM}: Visual expert for pretrained language models},
  author={Wang, Weihan and Lv, Qingsong and Yu, Wenmeng and Hong, Wenyi and Qi, Ji and Wang, Yan and Ji, Junhui and Yang, Zhuoyi and Zhao, Lei and XiXuan, Song and others},
  booktitle=NIPS,
  year={2024}
}

@article{qwen2.5,
  title={{Qwen2.5-VL} technical report},
  author={Bai, Shuai and Chen, Keqin and Liu, Xuejing and Wang, Jialin and Ge, Wenbin and Song, Sibo and Dang, Kai and Wang, Peng and Wang, Shijie and Tang, Jun and others},
  journal={arXiv preprint arXiv:2502.13923},
  year={2025}
}

@inproceedings{adam,
  title={Adam: A method for stochastic optimization},
  author={Kingma, Diederik P},
  booktitle=ICLR,
  year={2015}
}

@inproceedings{ho2022video,
  title={Video diffusion models},
  author={Ho, Jonathan and Salimans, Tim and Gritsenko, Alexey and Chan, William and Norouzi, Mohammad and Fleet, David J},
  booktitle=NIPS,
  year={2022}
}

@inproceedings{blattmann2023align,
  title={Align your latents: High-resolution video synthesis with latent diffusion models},
  author={Blattmann, Andreas and Rombach, Robin and Ling, Huan and Dockhorn, Tim and Kim, Seung Wook and Fidler, Sanja and Kreis, Karsten},
  booktitle=CVPR,
  year={2023}
}

@article{ho2022imagen,
  title={Imagen video: High definition video generation with diffusion models},
  author={Ho, Jonathan and Chan, William and Saharia, Chitwan and Whang, Jay and Gao, Ruiqi and Gritsenko, Alexey and Kingma, Diederik P and Poole, Ben and Norouzi, Mohammad and Fleet, David J and others},
  journal={arXiv preprint arXiv:2210.02303},
  year={2022}
}

@article{singer2022make,
  title={{Make-A-Video}: Text-to-video generation without text-video data},
  author={Singer, Uriel and Polyak, Adam and Hayes, Thomas and Yin, Xi and An, Jie and Zhang, Songyang and Hu, Qiyuan and Yang, Harry and Ashual, Oron and Gafni, Oran and others},
  journal={arXiv preprint arXiv:2209.14792},
  year={2022}
}

@inproceedings{chen2024diffusion,
  title={{Diffusion Forcing}: Next-token prediction meets full-sequence diffusion},
  author={Chen, Boyuan and Mart{\'\i} Mons{\'o}, Diego and Du, Yilun and Simchowitz, Max and Tedrake, Russ and Sitzmann, Vincent},
  booktitle=NIPS,
  year={2024}
}

@article{pont20172017,
  title={The 2017 davis challenge on video object segmentation},
  author={Pont-Tuset, Jordi and Perazzi, Federico and Caelles, Sergi and Arbel{\'a}ez, Pablo and Sorkine-Hornung, Alex and Van Gool, Luc},
  journal={arXiv preprint arXiv:1704.00675},
  year={2017}
}

@article{xu2018youtube,
  title={Youtube-vos: A large-scale video object segmentation benchmark},
  author={Xu, Ning and Yang, Linjie and Fan, Yuchen and Yue, Dingcheng and Liang, Yuchen and Yang, Jianchao and Huang, Thomas},
  journal={arXiv preprint arXiv:1809.03327},
  year={2018}
}

@article{wu2024longvideobench,
      title={LongVideoBench: A Benchmark for Long-context Interleaved Video-Language Understanding}, 
      author={Haoning Wu and Dongxu Li and Bei Chen and Junnan Li},
      journal={arXiv preprint arXiv:2407.15754},
      year={2024},
}

@inproceedings{fu2025mme,
  title={Mme: A comprehensive evaluation benchmark for multimodal large language models},
  author={Fu, Chaoyou and Chen, Peixian and Shen, Yunhang and Qin, Yulei and Zhang, Mengdan and Lin, Xu and Yang, Jinrui and Zheng, Xiawu and Li, Ke and Sun, Xing and others},
  booktitle=NIPS,
  year={2025}
}

@inproceedings{rombach2022high,
  title={High-resolution image synthesis with latent diffusion models},
  author={Rombach, Robin and Blattmann, Andreas and Lorenz, Dominik and Esser, Patrick and Ommer, Bj{\"o}rn},
  booktitle=CVPR,
  year={2022}
}

@inproceedings{mast3r_eccv24,
      title={Grounding Image Matching in 3D with MASt3R}, 
      author={Vincent Leroy and Yohann Cabon and Jerome Revaud},
      booktitle=ECCV,
      year = {2024}
}

@article{zhou2022magicvideo,
  title={Magicvideo: Efficient video generation with latent diffusion models},
  author={Zhou, Daquan and Wang, Weimin and Yan, Hanshu and Lv, Weiwei and Zhu, Yizhe and Feng, Jiashi},
  journal={arXiv preprint arXiv:2211.11018},
  year={2022}
}

@inproceedings{wu2023tune,
  title={Tune-a-video: One-shot tuning of image diffusion models for text-to-video generation},
  author={Wu, Jay Zhangjie and Ge, Yixiao and Wang, Xintao and Lei, Stan Weixian and Gu, Yuchao and Shi, Yufei and Hsu, Wynne and Shan, Ying and Qie, Xiaohu and Shou, Mike Zheng},
  booktitle=ICCV,
  year={2023}
}

@inproceedings{chen2024videocrafter2,
  title={Videocrafter2: Overcoming data limitations for high-quality video diffusion models},
  author={Chen, Haoxin and Zhang, Yong and Cun, Xiaodong and Xia, Menghan and Wang, Xintao and Weng, Chao and Shan, Ying},
  booktitle=CVPR,
  year={2024}
}

@inproceedings{zi2025cococo,
  title={Cococo: Improving text-guided video inpainting for better consistency, controllability and compatibility},
  author={Zi, Bojia and Zhao, Shihao and Qi, Xianbiao and Wang, Jianan and Shi, Yukai and Chen, Qianyu and Liang, Bin and Xiao, Rong and Wong, Kam-Fai and Zhang, Lei},
  booktitle=AAAI,
  year={2025}
}

@article{liu2025rolling,
  title={Rolling forcing: Autoregressive long video diffusion in real time},
  author={Liu, Kunhao and Hu, Wenbo and Xu, Jiale and Shan, Ying and Lu, Shijian},
  journal={arXiv preprint arXiv:2509.25161},
  year={2025}
}

@inproceedings{alayrac2022flamingo,
  title={Flamingo: a visual language model for few-shot learning},
  author={Alayrac, Jean-Baptiste and Donahue, Jeff and Luc, Pauline and Miech, Antoine and Barr, Iain and Hasson, Yana and Lenc, Karel and Mensch, Arthur and Millican, Katherine and Reynolds, Malcolm and others},
  booktitle=NIPS,
  year={2022}
}

@article{ye2023mplug,
  title={mplug-owl: Modularization empowers large language models with multimodality},
  author={Ye, Qinghao and Xu, Haiyang and Xu, Guohai and Ye, Jiabo and Yan, Ming and Zhou, Yiyang and Wang, Junyang and Hu, Anwen and Shi, Pengcheng and Shi, Yaya and others},
  journal={arXiv preprint arXiv:2304.14178},
  year={2023}
}

@article{animatediff,
  title={Animatediff: Animate your personalized text-to-image diffusion models without specific tuning},
  author={Guo, Yuwei and Yang, Ceyuan and Rao, Anyi and Liang, Zhengyang and Wang, Yaohui and Qiao, Yu and Agrawala, Maneesh and Lin, Dahua and Dai, Bo},
  journal={arXiv preprint arXiv:2307.04725},
  year={2023}
}

@inproceedings{edstedt2024roma,
title={{RoMa: Robust Dense Feature Matching}},
author={Edstedt, Johan and Sun, Qiyu and Bökman, Georg and Wadenbäck, Mårten and Felsberg, Michael},
booktitle={CVPR},
year={2024}
}

@inproceedings{chen2024restoreagent,
 author = {Chen, Haoyu and Li, Wenbo and Gu, Jinjin and Ren, Jingjing and Chen, Sixiang and Ye, Tian and Pei, Renjing and Zhou, Kaiwen and Song, Fenglong and Zhu, Lei},
 booktitle = {Advances in Neural Information Processing Systems},
 title = {RestoreAgent: Autonomous Image Restoration Agent via Multimodal Large Language Models},
 booktitle=NIPS,
 year = {2024}
}

@inproceedings{zhu2025agenticir,
  title = {An Intelligent Agentic System for Complex Image Restoration Problems},
  author = {Zhu, Kaiwen and Gu, Jinjin and You, Zhiyuan and Qiao, Yu and Dong, Chao},
  booktitle=ICLR,
  year = {2025},
}

@article{gu2025agenticposition,
  title = {Position: Agentic Systems Constitute a Key Component of Next-Generation Intelligent Image Processing},
  author = {Gu, Jinjin},
  journal={arXiv preprint arXiv:2505.16007},
  year = {2025},
}

@inproceedings{you2024depictqa,
  title = {Depicting Beyond Scores: Advancing Image Quality Assessment through Multi-modal Language Models},
  author = {You, Zhiyuan and Li, Zheyuan and Gu, Jinjin and Yin, Zhenfei and Xue, Tianfan and Dong, Chao},
  booktitle=ECCV,
  year = {2024},
}

@article{you2025edqa,
  title = {Enhancing Descriptive Image Quality Assessment With a Large-Scale Multi-Modal Dataset},
  author = {You, Zhiyuan and Gu, Jinjin and Cai, Xin and Li, Zheyuan and Zhu, Kaiwen and Dong, Chao and Xue, Tianfan},
  journal = TIP,
  year = {2025},
}

@inproceedings{you2026photoframer,
  title = {{PhotoFramer}: Multi-modal Image Composition Instruction},
  author = {You, Zhiyuan and Wang, Ke and Zhang, He and Cai, Xin and Gu, Jinjin and Xue, Tianfan and Dong, Chao and Zhang, Zhoutong},
  booktitle = CVPR,
  year = {2026},
}

@inproceedings{gu2020pipal,
  title = {{PIPAL}: A Large-Scale Image Quality Assessment Dataset for Perceptual Image Restoration},
  author = {Gu, Jinjin and Cai, Haoming and Chen, Haoyu and Ye, Xiaoxing and Ren, Jimmy S. and Dong, Chao},
  booktitle = ECCV,
  year = {2020},
}
}

\end{document}